\documentclass[a4paper,fleqn]{cas-dc}

\usepackage[super, numbers, compress]{natbib}
\usepackage{subfiles}
\usepackage{graphicx}
\usepackage{times}
\usepackage{epsfig}
\usepackage{graphicx}
\usepackage{amsmath}
\usepackage{amssymb}
\usepackage{booktabs}
\usepackage{bm}
\usepackage{multirow}
\usepackage{multicol}
\usepackage{flushend}
\usepackage{float}
\usepackage{stfloats}
\usepackage{subfiles}
\usepackage{color}
\usepackage{subfig}
\usepackage{colortbl} % color for table lines
\newcommand{\ie}{\textit{i}.\textit{e}.}
\newcommand{\eg}{\textit{e}.\textit{g}.}
\usepackage{dsfont}
\usepackage{bm}
\usepackage{bbm}
\usepackage{amstext}
\usepackage{amsfonts}
\usepackage{tabulary}
\usepackage{xurl}

\def\tsc#1{\csdef{#1}{\textsc{\lowercase{#1}}\xspace}}
\tsc{WGM}
\tsc{QE}
\begin{document}
\let\WriteBookmarks\relax
\def\floatpagepagefraction{1}
\def\textpagefraction{.001}

\shorttitle{Bodily expressed emotion understanding through integrating Laban Movement Analysis}    

\shortauthors{C. Wu et al.}  

\title [mode = title]{Bodily expressed emotion understanding through integrating Laban movement analysis}  

\author[1]{Chenyan Wu}[orcid=]\ead{czw390@psu.edu}
\author[1]{Dolzodmaa Davaasuren}[orcid=0000-0002-2810-902X]\ead{dud240@psu.edu}
\author[2]{Tal Shafir}\ead{tshafir1@univ.haifa.ac.il}
\author[3]{Rachelle Tsachor}[orcid=0000-0002-9640-2770]\ead{rtsachor@uic.edu}
\author[1,4]{James Z. Wang}[orcid=0000-0003-4379-4173]\cormark[1]\ead{jwang@ist.psu.edu}

\cortext[1]{Correspondence: jwang@ist.psu.edu}

\fnmark[1]

\credit{<Credit authorship details>}

\affiliation[1]{organization={Data Science and Artificial Intelligence Area, College of Information Sciences and Technology, The Pennsylvania State University}, city= {University Park}, state = {PA}, postcode = {16802}, country = {USA}}

\affiliation[2]{organization={The Emili Sagol Creative Arts Therapies Research Center, University of Haifa}, city= {Haifa}, country = {Israel}}

\affiliation[3]{organization={School of Theatre and Music, University of Illinois}, city= {Chicago}, state = {IL}, postcode = {60607}, country = {USA}}

\affiliation[4]{organization={Human-Computer Interaction Area, College of Information Sciences and Technology, The Pennsylvania State University}, city= {University Park}, state = {PA}, postcode = {16802}, country = {USA}}

\fntext[1]{Lead contact}

\begin{abstract}
Body movements carry important information about a person's emotions or mental state and are essential in daily communication. Enhancing the ability of machines to understand emotions expressed through body language can improve the communication of assistive robots with children and elderly users, provide psychiatric professionals with quantitative diagnostic and prognostic assistance, and aid law enforcement in identifying deception. 
This study develops a high-quality human motor element dataset based on the Laban Movement Analysis movement coding system and utilizes that to jointly learn about motor elements and emotions. Our long-term ambition is to integrate knowledge from computing, psychology, and performing arts to enable automated understanding and analysis of emotion and mental state through body language. This work serves as a launchpad for further research into recognizing emotions through analysis of human movement. 
\end{abstract}

\begin{highlights}
\item Body Motor Elements (BoME) dataset characterized by Laban Movement Analysis
\item Estimating Laban Movement Analysis motor elements with deep neural networks 
\item Jointly training dual-branch, dual-task Movement Analysis Network (MANet)
\item State-of-the-art performance on both motor element and emotion recognition tasks 
\end{highlights}


% \begin{graphicalabstract}
% \fbox{\includegraphics[width=0.9\textwidth]{images/GraphicalAbstract.pdf}}
% \end{graphicalabstract}

\maketitle

\section{Summary}
Bodily Expressed Emotion Understanding (BEEU) aims at automatically recognizing human emotional expressions from body movements. Psychological research has demonstrated that people often move using specific motor elements to convey emotions. This work takes three steps to integrate human motor elements to study BEEU. First, we introduce BoME (Body Motor Elements), a highly precise dataset for human motor elements. Secondly, we apply baseline models to estimate these elements on BoME, showing that deep learning methods are capable of learning effective representations of human movement. Finally, we propose a dual-source solution to enhance the BEEU model with the BoME dataset, which trains with both motor element and emotion labels and simultaneously produces predictions for both. Through experiments on the BoLD in-the-wild emotion understanding benchmark, we demonstrate the significant benefit of our approach. These results may inspire further research that utilizes human motor elements for emotion understanding and mental health analysis.

\section{Introduction}\label{Introduction}
Recognizing human emotional expressions from images or videos is a fundamental area of research in affective computing and computer vision, with numerous applications in robotics and human-computer interaction.~\citep{marcos2021emotional,cowie2001emotion} 
With the development of the Body Language Dataset (BoLD), a large-scale in-the-wild dataset for bodily expressed emotion, and the corresponding benchmark deep neural network models,\citep{luo2020arbee} research on emotion recognition has increasingly focused on bodily expressed emotion understanding (BEEU).\citep{bhattacharya2020step,filntisis2020emotion,huang2021emotion} In contrast to the well-studied problem of facial expression recognition,\citep{eleftheriadis2015multi,fabian2016emotionet,kollias2020deep,ruan2022adaptive} BEEU aims to automatically recognize emotional expressions from body movements.

Emotion recognition based on body movements has several benefits compared to facial inputs. Firstly, in crowded scenes, a person's facial area may be obscured or lack sufficient resolution, but body movements and postures can still be reliably detected. Secondly, research has shown that the body may be more diagnostic than the face for emotion recognition.\citep{aviezer2012body} Thirdly, facial areas may be inaccessible in some applications for privacy and confidentiality reasons. Fourthly, it is difficult to fake subtle emotions through body movements, whereas facial expressions can often be faked. Finally, using body movements as an additional modality can lead to more accurate recognition compared to using only facial images or videos.

BEEU presents a challenging task for artificial emotional intelligence due to several factors.\citep{wang2022emotion} 
Firstly, it is difficult to collect a large, accurately labeled affective dataset for training deep learning models because emotion interpretation is highly subjective.\citep{luo2020arbee,ye2017probabilistic} Secondly, the context, such as surrounding information and activities, must be considered to correctly interpret a person's bodily expressed emotion. Thirdly, emotion interpretation may depend on the demographics of both the observed person and the observer. Fourthly, the concept of emotion is not well-defined in psychology, making it difficult to develop a specific algorithm to detect a specific emotion. As a result, the gap between low-level pixel information and high-level emotion labels is too large for straightforward data-driven approaches, and it is necessary to develop intermediate representations to assist AI in bridging this gap.

In facial expression recognition, some studies use the facial action coding system (FACS) as an intermediate representation. They first detect Action Units (AUs), which are defined as the movements of certain facial muscles in FACS, and then recognize emotions based on these detections. This approach is motivated by the fact that people contract specific muscles (AUs) to produce particular facial expressions (\eg the corrugator muscle contracts to frown, expressing anger).

\begin{figure}[t!]
        \centering
        \includegraphics[width=0.98\linewidth]{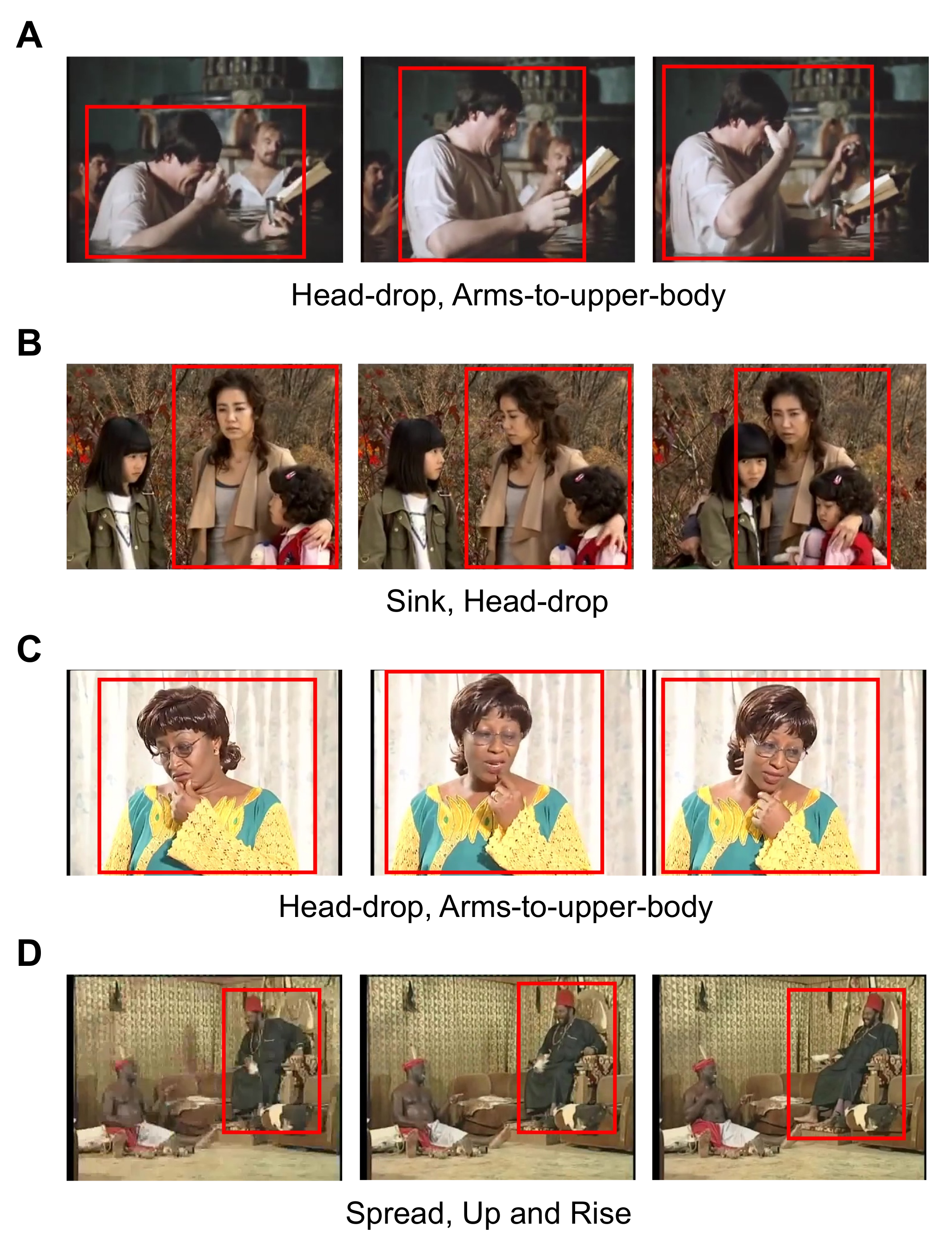}
        \caption{\textbf{Example video clips from the BoME dataset}\hfill\break
        Three sample frames are shown for each clip. Instances of interest are bounded by red boxes.The LMA motor elements annotated for each clip are shown.}
        \label{fig:example}
    \end{figure}
In the same way, we use specific body muscles and parts of the skeleton to perform actions that communicate our emotions. For example, people may touch their heads with their hands when expressing sadness, as shown in the first example in Figure~\ref{fig:example}.
Just as FACS does in facial expression recognition, we can describe specific movements that are common to all humans using the motor elements that make up those movements and then identify the relationship between those motor elements and bodily expressed emotion.

Motor elements are often more easily detectable in a video of a person compared to facial muscle movement. Additionally, these motor elements often have a clear definition, making them easy for AI to recognize. As a result, motor elements can serve as a suitable intermediate representation for BEEU, bridging the gap between low-level movement features and emotion category labels.

However, previous BEEU studies have rarely incorporated motor elements into their approaches.
\citeauthor{luo2020arbee}~\cite{luo2020arbee} only used low-level movement features such as speed and acceleration of various joints.
Other approaches~\citep{bhattacharya2020step,filntisis2020emotion,huang2021emotion} utilize techniques developed for video or action recognition, directly inputting human movement videos into a video-recognition network and predicting emotion categories without considering motor element understanding.

In this work, we present a new paradigm for BEEU that incorporates motor element analysis. Specifically, we use deep neural networks to recognize motor elements and then use them as intermediate features to recognize emotion.

To achieve this goal, the first challenge is to acquire a motor elements dataset.
Currently, there are no suitable public datasets that we can use because deep learning has not been widely utilized in prior studies on human motor elements.
Therefore, we created the BoME (Body Motor Elements) dataset, consisting of 1,600 clips of human movements with high-precision, expert-supplied movement labels.
We use the AVA video dataset~\citep{gu2018ava} as the video source and applied the Laban Movement Analysis (LMA) system to describe the motor elements.
LMA, which was first developed in the early 20th century in the dance community, is now an internationally recognized framework for describing and comprehending human bodily motions. It characterizes human movements into five categories: {\bf Body}, {\bf Effort}, {\bf Shape}, {\bf Phrasing}, and {\bf Space}, and includes about a hundred detailed elements.
Based on the preliminary psychological research on the relationship between LMA and emotion,\citep{shafir2016emotion,melzer2019we,van2021move} we selected eleven emotion-related LMA elements to focus the annotation effort.
We designed a procedure for collecting the dataset and invited a Certified Movement Analyst (CMA), an expert in LMA, to annotate whether the LMA elements are present in the human movement clips. Some examples of the BoME dataset are shown in Figure~\ref{fig:example}. Further details on the BoME dataset can be found in the Experimental Procedures section.

Using the established BoME dataset, we examine whether deep neural networks can learn an effective representation of human movement. We deploy several state-of-the-art video recognition networks on BoME to estimate the LMA elements and investigate the impact of factors such as the video sampling rate and pre-training datasets on network performance. The results show that these methods, particularly the Video Swin Transformer~\cite{liu2022video}, perform well on the BoME dataset, indicating that deep neural networks can learn a suitable movement representation from BoME.

Finally, we conduct experiments to enhance BEEU by using the BoME dataset as an additional source of supervision.
We design a dual-branch, dual-task network called Movement Analysis Network (MANet), whose branches produce predictions for bodily expressed emotion and LMA labels, respectively.
To effectively utilize the movement representation in emotion recognition, we merge the LMA branch features into the emotion branch.
We also propose a new Bridge loss that allows the LMA prediction to supervise the emotion prediction.
By using a weak supervision strategy, we jointly train MANet on the BEEU benchmark BoLD and the BoME dataset.
The BEEU results on the BoLD validation and test sets show that our approach significantly outperforms all the single-task (\ie, only BEEU) baselines.
\section{Results}\label{section:results}
\subsection{Statistical analysis confirms the effectiveness of the LMA motor elements}
\newcolumntype{S}{>{\arraybackslash}m{10cm}}
\definecolor{light-gray}{gray}{0.6}
\begin{table*}[ht!]
    \begin{center}
    \setlength{\tabcolsep}{5pt}
    \caption{\textbf{The eleven LMA elements coded in the BoME dataset}\hfill\break Psychological studies indicate that the first four elements are associated with sadness and the rest with happiness.\hfill\break $*$ Up and Rise are two elements within the Shape and Space categories, respectively. We follow \citeauthor{shafir2016emotion,melzer2019we}~\cite{shafir2016emotion,melzer2019we} to merge them into one element because their movements are similar.}\label{table:list}
    \label{table:anchor}
    \begin{tabular}{p{0.4cm}llS}
    \toprule
     & LMA Element & LMA Category & Description \\
    \hline
    \multirow{7}{*}{\rotatebox[origin=c]{90}{Sadness}} & Passive Weight &      Effort         &  {\footnotesize Lack of active attitude towards weight, resulting in sagging, heaviness, limpness, or dropping.} \\
    \arrayrulecolor{light-gray}\cline{2-4}\arrayrulecolor{black}
         &  Arms-to-upper-body   &     Body          & {\footnotesize Hands or arms touching any part of the upper body (head, neck, shoulders, or chest).} \\
    \arrayrulecolor{light-gray}\cline{2-4}\arrayrulecolor{black}
         &     Sink       &       Shape        & {\footnotesize Shortening of the torso and head and letting the center of gravity drop downwards, so the torso is convex on the front.} \\
    \arrayrulecolor{light-gray}\cline{2-4}\arrayrulecolor{black}
         &   Head-drop      &       Body        & {\footnotesize Releasing the weight of the head forward and downward, using the quality of Passive Weight. Dropping the head down.} \\
    \hline
    \multirow{11}{*}{\rotatebox[origin=c]{90}{Happiness}} &      Jump       &      Body         & {\footnotesize Any type of jumping.}  \\
    \arrayrulecolor{light-gray}\cline{2-4}\arrayrulecolor{black}
    &      Rhythmicity         &         Phrasing        & {\footnotesize Rhythmic repetition of any aspect of the movement, like bouncing, rocking, bobbing, twisting, from side to side, etc.}  \\
    \arrayrulecolor{light-gray}\cline{2-4}\arrayrulecolor{black}
    &      Spread              &          Shape       & {\footnotesize When the mover opens his body to become wider.} \\
    \arrayrulecolor{light-gray}\cline{2-4}\arrayrulecolor{black}
    &      Free Flow      &        Effort         & {\footnotesize Lessening movement control, moving like you ‘go with the flow’.} \\
    \arrayrulecolor{light-gray}\cline{2-4}\arrayrulecolor{black}
    &      Light Weight        &        Effort        & {\footnotesize Moving with a sense of lightness and buoyancy of the body or its parts; gentle or delicate movement with very little pressure and a sense of letting go upward.} \\
    \arrayrulecolor{light-gray}\cline{2-4}\arrayrulecolor{black}
    &      Up and Rise$^*$         &   Space/Shape  & {\footnotesize Up means going in the upward direction in the allocentric space. Rise means raising the chest up by lengthening the torso.} \\
    \arrayrulecolor{light-gray}\cline{2-4}\arrayrulecolor{black}
    &        Rotation          &         Shape        & {\footnotesize Rotating a body part, or turning the entire body around itself in space, like in a Sufi dance.} \\
    \bottomrule
    \end{tabular}
    \end{center}
    \end{table*}

\begin{figure}[t!]
    \centering
    \includegraphics[trim={0 0 0 0},clip,width=0.98\linewidth]{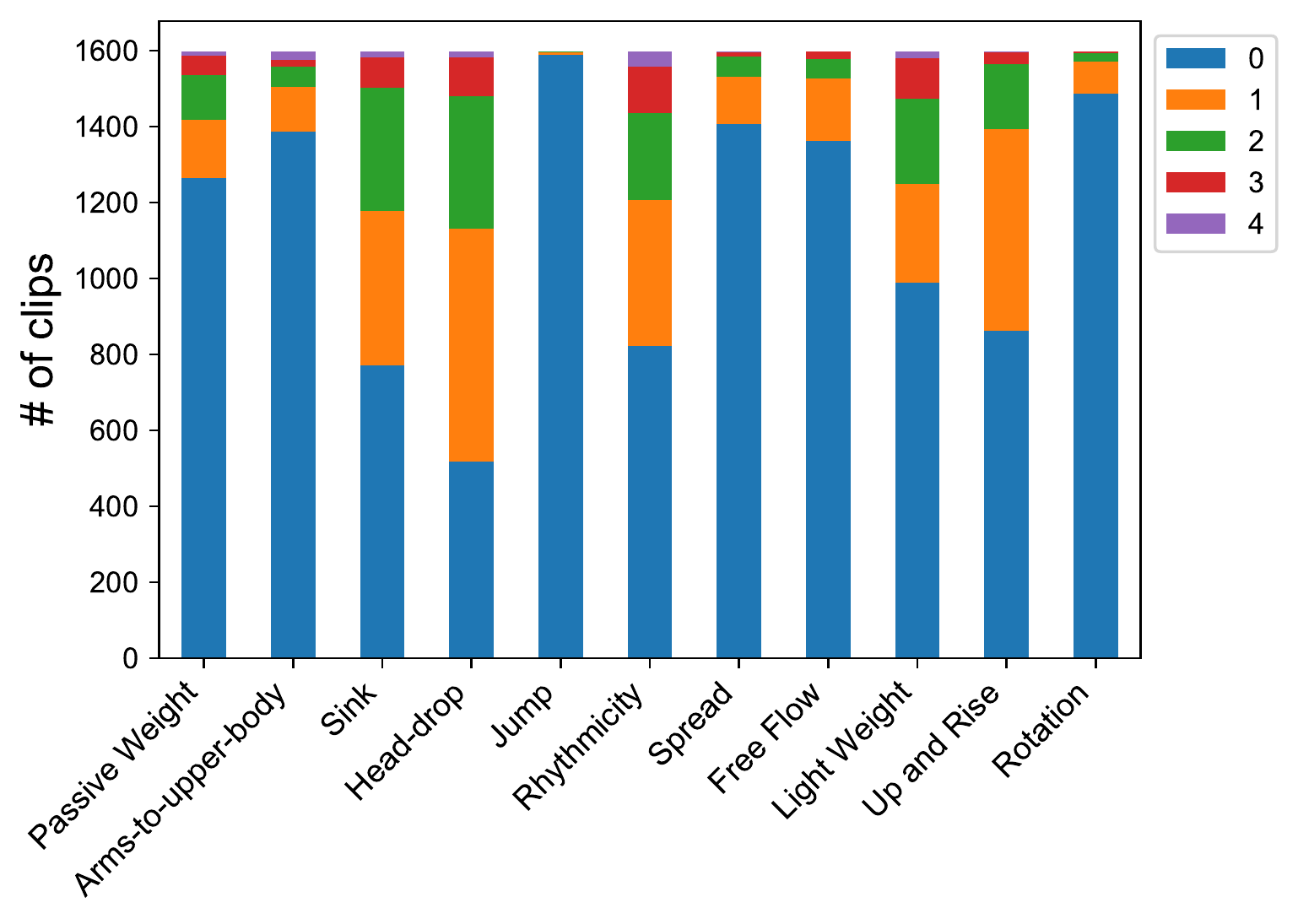}
    \caption{\textbf{Distribution of the five-level labels for each LMA element}}
    \label{fig:barplot}
\end{figure}
In order to improve the ability of deep neural networks to learn human movement representation and subsequently enhance emotion recognition, we developed a high-precision motor element dataset named BoME. This dataset consists of 1,600 human video clips, each of which has been meticulously annotated with LMA labels.
To achieve a balance between precision and utility, we annotated each clip with eleven LMA elements.
Research has indicated that these elements are associated with sadness and happiness and are relevant for emotion elicitation and emotional expression, making them useful for understanding bodily expression.~\cite{shafir2016emotion,melzer2019we,van2021move}
Additionally, annotating eleven elements is not a particularly laborious task for LMA experts, ensuring the quality of the dataset.
Table~\ref{table:list} lists the eleven elements and their associated emotions, LMA categories, and descriptions.

For each LMA element, we assigned a five-level label based on the duration and intensity of the element in the clip, with level 0 indicating no presence and level 4 indicating maximum presence. The distribution of the five levels for each LMA element is shown in Figure~\ref{fig:barplot}.
We can see the Head-drop element is relatively commonly present in videos, while Jump cases are rare.
The five-level label can also be treated as a binary label, with level 0 representing a negative label and non-zero levels representing a positive label.
On average, each clip includes 3.2 positive LMA elements, with a minimum of one and a maximum of ten positive elements per clip in the BoME dataset.

\begin{figure}[t!]
    \centering
    \includegraphics[trim={0 0 0 0},clip,width=0.98\linewidth]{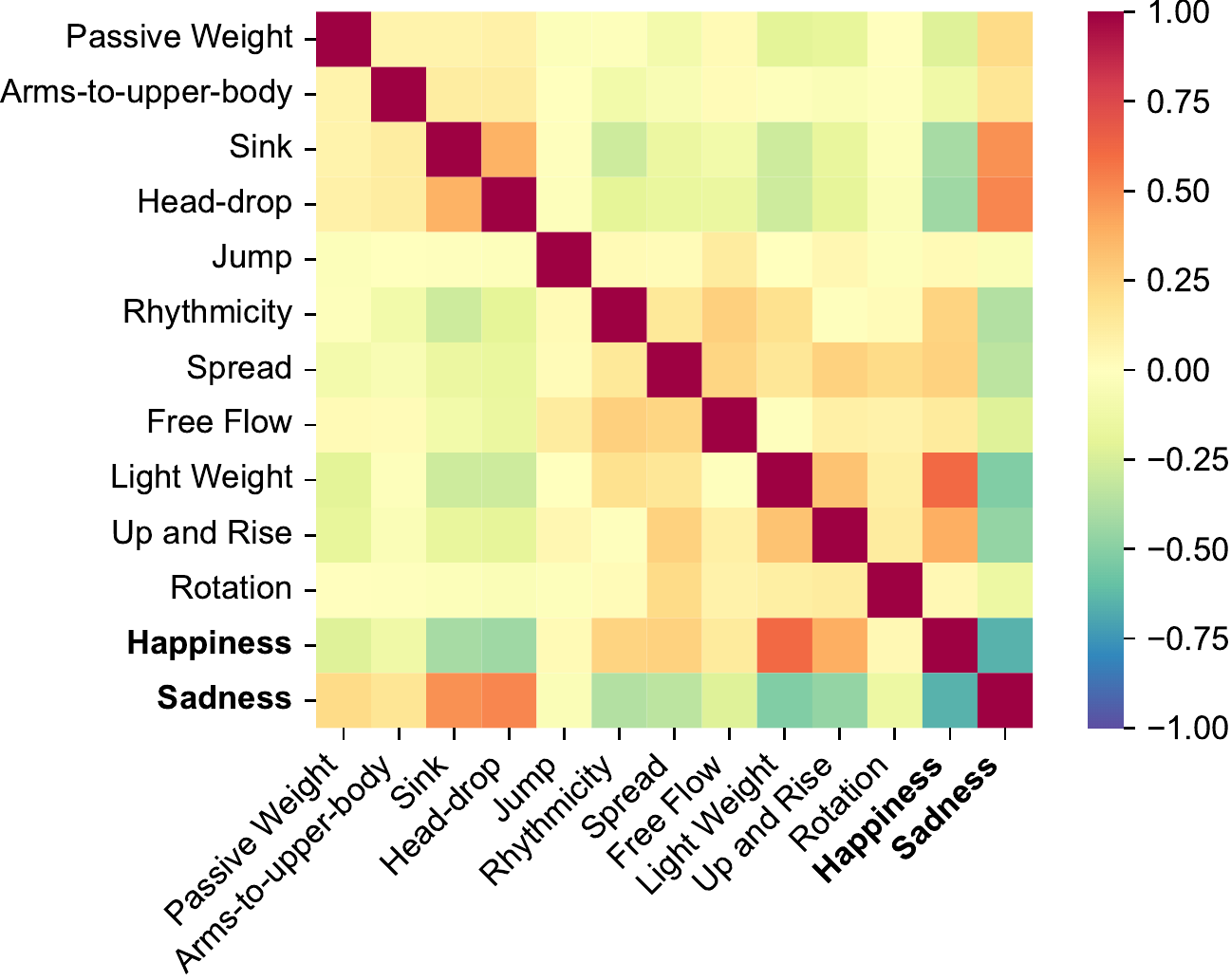}
    \caption{\textbf{Correlation among LMA elements and emotions}\hfill\break Emotion names are written in \textbf{bold}.}
    \label{fig:lma_corr}
    \end{figure}

To confirm the association of the LMA element labels in BoME with specific emotions, the annotators added an emotion label to each clip, which was limited to three categories: happiness, sadness, and other emotions. 
While emotion recognition studies often require a greater number of emotional categories, we limited ourselves to these three categories in order to verify the effectiveness of the LMA element labels.

For each human clip in BoME, we assigned a value of 0 or 1 to the happiness and sadness emotion categories, respectively, based on the emotion label. 
We also had the five-level label (ranging from 0 to 4) for each LMA element.
Using these values, we calculated the correlation between the LMA elements and the two emotion categories (happiness and sadness) across the entire dataset, as shown in Figure~\ref{fig:lma_corr}.
We found that Sink and Head-drop were strongly positively correlated with sadness, while Light Weight, Up, and Rise were notably positively correlated with happiness. These findings are consistent with previous psychological studies.~\cite{shafir2016emotion,melzer2019we,van2021move}
On the other hand, Jump and Rotation did not show a significant correlation with either happiness or sadness. This may be due to the relatively small sample size for these elements. Additionally, based on the emotion labels, we found that instances of sadness accounted for 59.9\% of the BoME dataset, while cases of happiness accounted for only 22.2\%. This may be because sadness is more easily recognizable to human annotators than other emotional categories.

\subsection{Deep neural networks are capable of estimating LMA motor elements}

To investigate the ability of deep learning-based methods to learn an effective human movement representation, we attempted to use deep neural networks to estimate the LMA elements on the BoME dataset.
We randomly divided the BoME dataset into a training set with 1,448 samples and a test set with 152 samples. 
The original LMA element labels had five levels, but as shown in Figure~\ref{fig:barplot}, levels 3 and 4 had relatively small sample sizes, which made it challenging for models to estimate. Additionally, deep neural networks may not be able to judge the duration of each element in a clip as accurately as LMA experts.
To simplify the task, we treated the LMA element estimation problem as a multi-label binary classification problem, where level 0 was assigned as a negative label and non-zero levels were treated as positive labels.

We evaluated the classification performance using two metrics: average precision (AP, \ie, the area under the precision-recall curve) and the area under the receiver operating characteristic curve (AUC-ROC).
We report the mean average precision (mAP) and mean AUC-ROC (mAR) over all categories of LMA elements.
It is worth noting that we only used ten elements, excluding the Jump element, as the dataset contained too few (11 total) samples for Jump.

As a natural first attempt, we utilize a range of video recognition algorithms for the estimation of LMA elements from videos. These algorithms can be broadly classified into two categories based on the input modality: RGB-based and skeleton-based. The RGB-based algorithms can further be divided into 2D convolution networks, 3D convolution networks, and Transformer-based networks. 
As a representative of each type, we have selected the Temporal Segment Network (TSN)~\cite{wang2018temporal} for 2D convolution networks, SlowFast~\cite{feichtenhofer2019slowfast} for 3D convolution networks, and Video Swin Transformer~\cite{liu2022video} (referred to as V-Swin in the following) for Transformer-based networks.
For the skeleton-based algorithm, we have adopted the state-of-the-art algorithm  PoseC3D.~\cite{duan2022revisiting}
In order to apply these algorithms to the BoME dataset, we have made the necessary modifications, including cropping the region containing the human from the video as input. We have largely followed the original implementation settings of the publicly available codebase MMaction2~\cite{2020mmaction2} in our experiments. Additional descriptions of the four algorithms and implementation details can be found in the Experimental Procedures section.

\begin{table*}[ht!]
    \caption{\textbf{Benchmarking the BoME dataset with four deep learning-based methods}\\
    ``Samples'' means number of temporal clip $\times$ number of frames per clip in \textbf{training} when sampling one long video. ``Views'' indicates number of temporal clip $\times$ number of spatial crop during \textbf{inference}. 
    }
    \centering
    \begin{tabular}{lcccccccc}
    \toprule
    Method & Modality & Pretrain & mAP(\%) & mAR(\%) & Samples &  Views & FLOPs ($\times 10^{9}$) & Param. ($\times 10^{6}$) \\
    \hline
    TSN  & RGB & Scratch  & 40.71 & 60.42  & 8 × 1 &  25 × 3 & 7.3 & 23.5 \\
    TSN  & RGB & ImageNet-1k  & 46.43 & 66.90  & 8 × 1 &  25 × 3 & 7.3 & 23.5 \\
    TSN  & RGB & Kinetics-400 & 49.03 & 68.58  & 8 × 1 &  25 × 3 &  7.3 & 23.5   \\
    TSN  & RGB & Kinetics-400 & \textbf{51.02} & 69.91  & 16 × 1 &  25 × 3 &  7.3 & 23.5   \\
    TSN  & RGB & Kinetics-400 & 50.58 & \textbf{70.35}  & 24 × 1 &  25 × 3 &  7.3 & 23.5   \\
    \hline
    SlowFast & RGB & Scratch  & 39.20 & 59.28  & 1 × 32 &  10 × 3 &  174 & 62.0  \\
    SlowFast & RGB & Kinetics-400 & \textbf{49.27} & \textbf{69.23}  & 1 × 32 &  10 × 3 &  174 & 62.0 \\
    SlowFast & RGB & Kinetics-400 & 48.62 & 68.19  & 1 × 48 &  10 × 3 &  260 & 62.0 \\
    SlowFast & RGB & Kinetics-400 & 47.70 & 68.12  & 1 × 64 &  10 × 3 &  347 & 62.0 \\
    \hline
    V-Swin & RGB & Scratch    & 39.58 & 59.33 &  1 × 32 &  4 × 3  & 282 & 88.1 \\
    V-Swin & RGB & ImageNet-1k    & 43.15 & 62.32 &  1 × 32 &  4 × 3  & 282 & 88.1 \\
    V-Swin & RGB & ImageNet-21k   & 48.88 & 66.49 &  1 × 32 &  4 × 3  &  282 & 88.1 \\
    V-Swin & RGB & Kinetics-400   & \textbf{53.67} & \textbf{72.82} & 1 × 32 &  4 × 3 & 282 & 88.1  \\
    V-Swin & RGB & Kinetics-400   & 50.78 & 69.55 &  1 × 48 &  4 × 3 & 423 & 88.1  \\
    V-Swin & RGB & Kinetics-400   & 46.79 & 64.42 &  1 × 64 &  4 × 3 & 564 & 88.1  \\
    \hline
    PoseC3D & Skeleton & Scratch            & 38.88 &  58.45  & 1 × 48 &  1 × 2 & 15 & 3.0 \\
    PoseC3D & Skeleton & Kinetics-400 & 44.75 &  61.75  & 1 × 48 &  1 × 2 & 15 & 3.0 \\
    PoseC3D & Skeleton & Kinetics-400 & \textbf{50.30} &  64.72  & 1 × 72 &  1 × 2 & 22 & 3.0 \\
    PoseC3D & Skeleton & Kinetics-400 & 46.93 &  \textbf{64.93}  & 1 × 96 &  1 × 2 & 29 & 3.0 \\
    \bottomrule
    \end{tabular}
    \label{tab:k400}
    \end{table*}

    \begin{figure*}[ht!]
        \centering
        \includegraphics[width=0.99\textwidth]{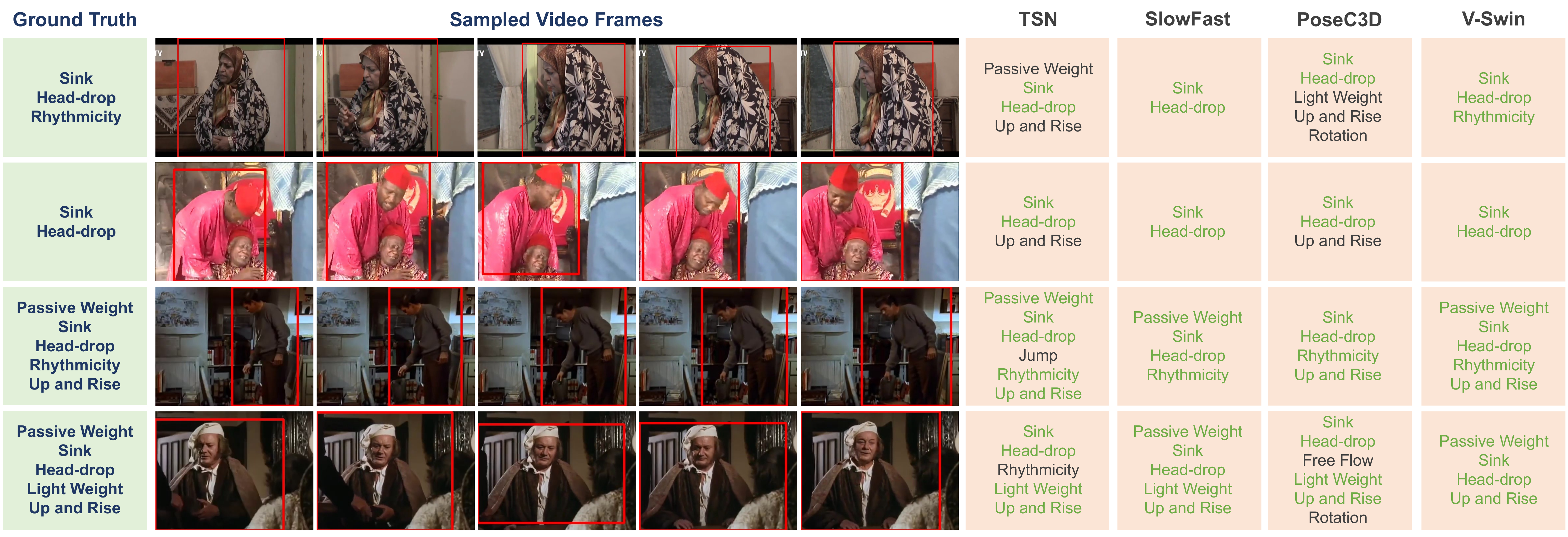}
        \caption{\textbf{Example LMA element estimation results on the BoME dataset}\hfill\break Five frames sampled from each clip are shown. The predicted LMA elements that are also in the ground truth list are shown in green color.}
        \label{fig:result_lma1}
        \end{figure*}

The performance of the four algorithms is compared in Table~\ref{tab:k400}. 
To optimize the estimation performance of each algorithm, we employed the pre-training technique and experimented with different video sample rates.
These methods can be trained from scratch or pre-trained using various existing datasets, such as the image classification dataset ImageNet-1K/ImageNet-22K~\cite{deng2009imagenet} and the video recognition dataset Keneicts400.~\cite{kay2017kinetics}
We followed the original algorithms' strategies for video sampling.
Specifically, TSN segments the full-length video into several clips, each including a single frame.
Regarding SlowFast and V-Swin, we sample one clip from the full-length video, where each clip contains several frames with a temporal stride of 2.
PoseC3D treats the full-length video as a single clip and uniformly samples certain frames from it.
The number of clips per video and the number of frames per clip are listed in Table~\ref{tab:k400}.

Our analysis of the results reveals the following insights. 
First, pre-training significantly enhances the performances of all four algorithms. 
Specifically, pre-training with ImageNet leads to notable improvement compared to training from scratch, with gains of 5.72 mAP(\%) and 6.48 mAR(\%) for TSN, and 3.57 mAP(\%) and 2.99 mAR(\%) for V-Swin. 
Pre-training on Kinetics-400 produces even more substantial improvements, with increases of 8.32, 10.07, 14.09, and 5.87 mAP(\%) for TSN, SlowFast, V-Swin, and PoseC3D, respectively. 
This is not surprising, as the Kinetics-400 dataset is specifically designed for human activity classification, and some human characteristics features can be transferred to the LMA estimation task.

Additionally, our results suggest that the sample rate at which the input video is extracted into frames may impact performance. Higher density of frame samples within a video may allow for more information to be extracted, but may also hinder the model's ability to analyze such packed frames, leading to a decrease in performance. 
TSN achieves the best performance when the video is split into 16 clips.
SlowFast and V-Swin exhibit worse performance with denser sampling rates than the default rate.
PoseC3D performs optimally while a sampling rate of 72 frames per video, the densest among the four algorithms.
This may be because PoseC3D uses skeleton coordinates as input, which may be easier for the neural network to interpret compared to images.

Finally, our results show that V-Swin achieves the best performance (53.67 mAP(\%) and 72.82 mAR(\%)) among the four algorithms, with 32 samples per video and pre-training on Kinetics-400. Despite using only skeleton input, PoseC3D obtains competitive mAP performance compared to TSN and SlowFast.

\begin{figure*}[ht!]
    \centering
    \includegraphics[trim=0 0 0 0, width=0.97\textwidth]{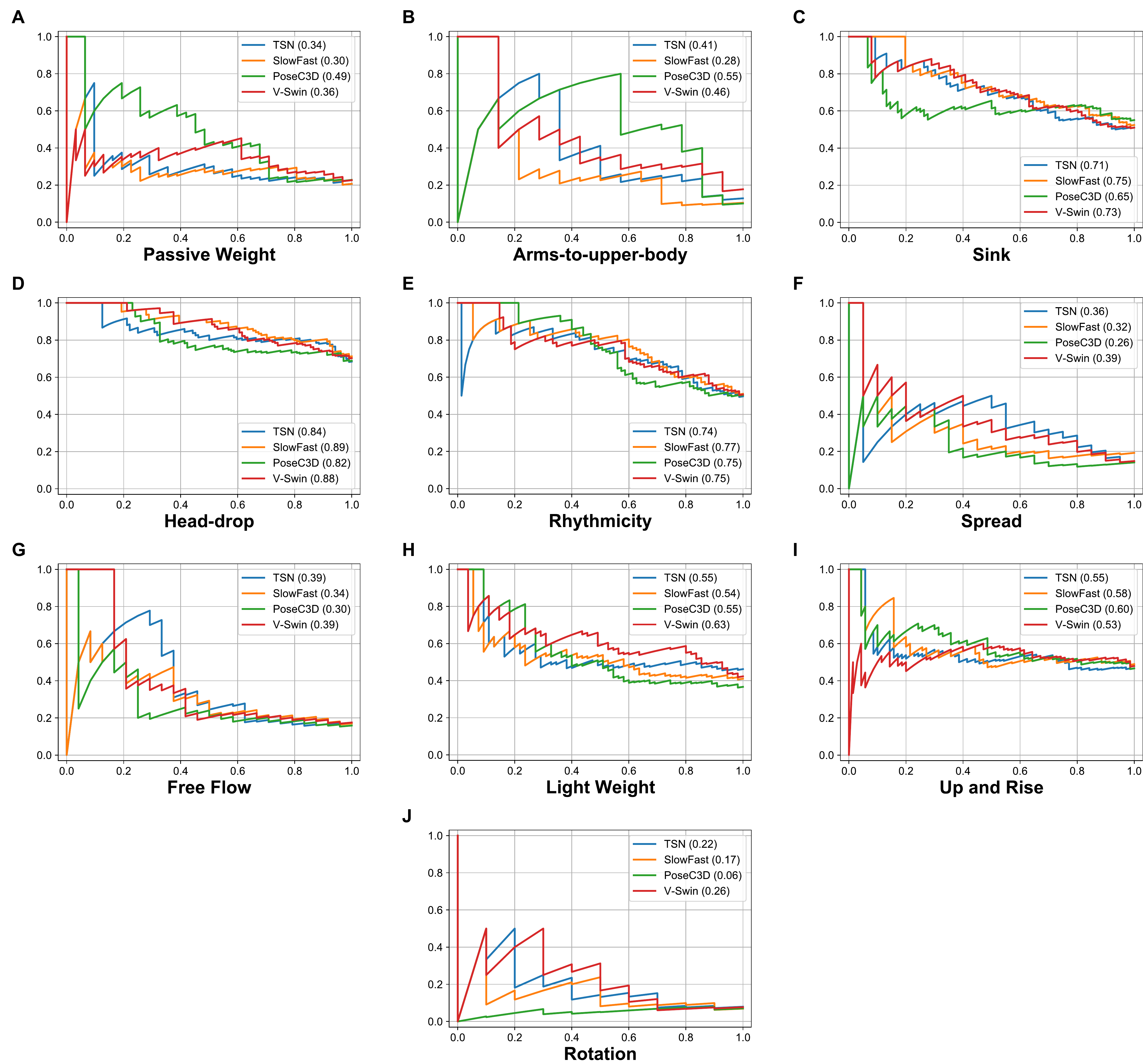}
    \caption{\textbf{Precision-recall curves for various models on ten LMA elements}\hfill\break
    The x-axis represents the recall, and the y-axis represents the precision. The AP (Average Precision) score of the corresponding model is indicated in parentheses after the model name in the legend.
    }
    \label{fig:ap}
    \end{figure*}

Based on the mAP values, we have selected the best-performing model for each algorithm. 
Some qualitative examples of LMA element estimation by different models, along with the ground truth, are shown in Figure~\ref{fig:result_lma1}.
We also conducted a breakdown analysis.
Figure~\ref{fig:ap} presents the precision-recall curves for various models over all the LMA elements. It is notable that V-Swin outperforms the other algorithms on the Light Weight and Rotation elements.
PoseC3D performs particularly well on the Passive Weight and Arms-to-upper-body elements, but poorly on the Spread and Rotation elements.

\begin{figure*}[ht!]
    \centering
    \includegraphics[width=0.95\textwidth]{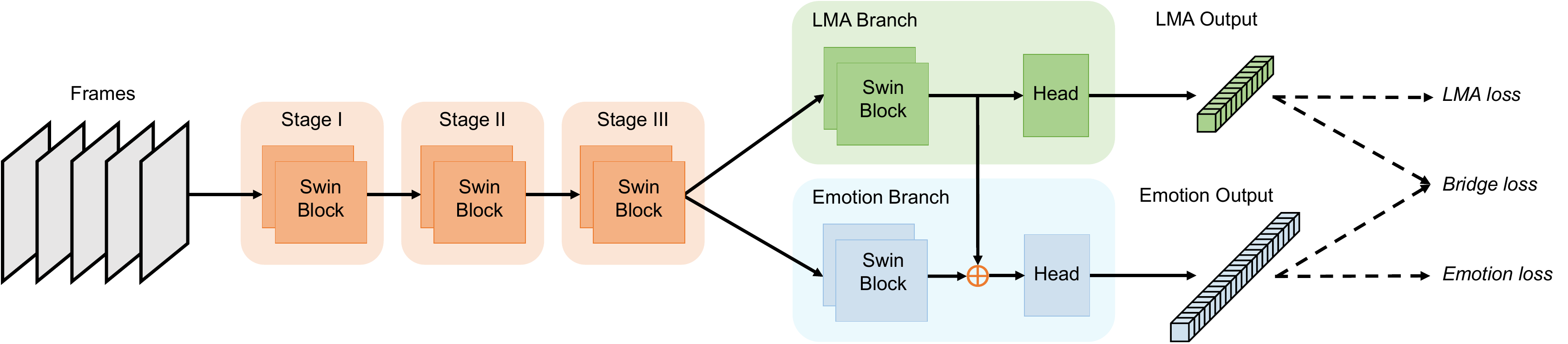}
    \caption{\textbf{The framework of the proposed Movement Analysis Network (MANet)}}
    \label{fig:lma}
    \end{figure*}

\subsection{LMA enhances bodily expressed emotion understanding}
In this subsection, we aim to accomplish the ultimate goal of enhancing BEEU by integrating LMA element labels.
As aforementioned, several phycological studies have demonstrated a strong correspondence between the eleven LMA elements and emotion categories sadness and happiness.\cite{shafir2016emotion,melzer2019we,van2021move}
Our previous statistical analysis has confirmed this finding in the BoME dataset.
Furthermore, we have shown that deep neural networks can learn an effective body movement representation from BoME. The optimal performance for LMA element estimation on BoME is 53.67 mAP(\%) and 72.72 mAR(\%), which is significantly higher than the best results achieved in BEEU (19.30 mAP(\%) and 66.94 mAR(\%)) on the bodily expression benchmark BoLD. This is likely due to the fact that LMA elements have a more objective definition than emotion categories, as the presence of LMA elements in a human video depends solely on the body movement, whereas emotion labels may also be influenced by the annotators' emotional state. In summary, emotion and LMA element labels are related, and LMA element labels are easier for deep neural networks to learn. Therefore, incorporating the human movement features learned from BoME into BEEU is a promising approach.

To achieve our goal of improving BEEU, we need to train and test on the BEEU benchmark dataset BoLD, using the BoME dataset as an additional training source. Therefore, in this set of experiments, we jointly train on the BoLD training set and the entire BoME dataset and then evaluate the model on the BoLD validation and test sets.

Figure~\ref{fig:lma} illustrates the proposed method, which involves the development of a dual-branch, dual-task network called MANet.
We have made three crucial implementations to allow the LMA annotations assist BEEU.
\textit{First}, we designed a dual-branch network structure, allowing the network to output LMA and emotion predictions simultaneously.
We used the first three stages of the V-Swin network~\cite{liu2022video} as the backbone of MANet, followed by the Emotion branch and the LMA branch, both of which consist of several Swin blocks.
Since predicting LMA and emotion are both multi-label binary classification tasks, we applied the multi-label binary cross-entropy loss to calculate the LMA loss and the Emotion loss.
\textit{Second}, we deployed a fusion operation by adding the LMA branch features to the emotion features, allowing the emotion features to be predicted with the assistance of human movement features.
\textit{Third}, we proposed the Bridge loss to connect the LMA and emotion prediction, based on the relationship between the LMA and certain emotion categories (\ie, happiness and sadness).
It is worth noting that we used a threshold $\epsilon$ in Bridge loss to control the extent to which the LMA prediction supervises the emotion prediction.

Furthermore, we utilize a weakly supervised training approach to facilitate joint training in cases where some BoLD samples are missing LMA labels and some BoME samples are missing emotion labels. Detailed information on the model structure, loss functions, and training procedure can be found in the Experimental Procedures section.

\begin{table}[ht!]
    \caption{\textbf{Ablation on the architecture and Bridge loss}\hfill\break Evaluation is done on the BoLD validation set.}\label{tab:loss}
    \centering
    \setlength{\tabcolsep}{4pt}
    \begin{tabular}{ccc cc}
    \toprule
    Dual-branch & Fusion  & Bridge Loss &  mAP(\%) & mAR(\%)   \\
    \hline
    -  & - & - & 19.97 & 67.16   \\
    \checkmark & - & - & 19.94 & 67.34   \\
    \checkmark & \checkmark  & - & \textbf{20.43} & \textbf{67.76}  \\
    \hline
    \checkmark & \checkmark & w/o $\epsilon$  & 20.42	& 66.88  \\
    \checkmark & \checkmark & $\epsilon=0.7$  & 19.78	& 67.44   \\
    \checkmark & \checkmark & $\epsilon=0.8$  & 20.55	& 67.43   \\
    \checkmark & \checkmark & $\epsilon=0.9$  & \textbf{21.25} & \textbf{68.32}   \\
    \checkmark & \checkmark & $\epsilon=0.99$ & 20.67	& 67.57 \\
    \bottomrule
    \end{tabular}
\end{table}

\begin{figure*}[ht!]
    \centering    
    \includegraphics[trim={0 0 8 0}, clip,width=0.99\textwidth]{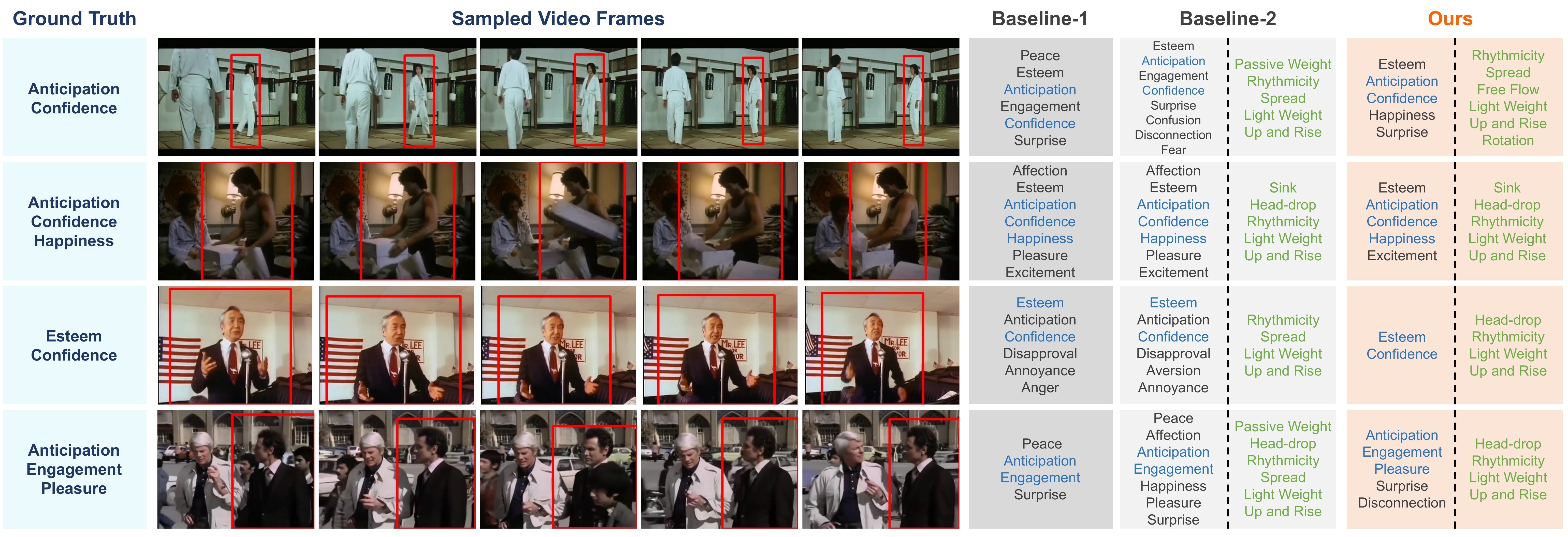}
    \caption{\textbf{Example bodily expressed emotion understanding results on the BoLD validation set}\hfill\break Categorical emotion labels are predicted based a video clip of a person. Five frames sampled from each clip are shown. The predicted emotion labels that are also in the ground truth list are shown in blue color. Baseline-1 is the original V-Swin without dual-branch and fusion. Baseline-2 is the MANet model without the Bridge loss. Ours is the final model of MANet. We also present the predicted LMA elements in green color. Baseline-1 does not provide LMA predictions because it only outputs emotion prediction.}
    \label{fig:result_lma2}
    \end{figure*}

Table~\ref{tab:loss} presents the results of the ablation study.
The first set of experiments evaluates the impact of the model structure on performance.
The method without the dual-branch and fusion components refers to only training emotion labels using the original V-Swin architecture. In this case, the performance is 19.97 mAP(\%) and 67.16 mAR(\%) on the BoLD validation set.
Incorporating the dual-branch structure, but omitting fusion, does not significantly improve performance compared to the original V-Swin.
However, the fusion operation operation leads to a 0.46 mAP(\%) and 0.60 mAR(\%) increase over the original V-Swin.
This suggests that multi-task training and feature fusion are both necessary for improving BEEU.
The effectiveness of the Bridge loss is also analyzed.
As detailed in Experimental Procedures, the initial version of the Bridge loss does not incorporate the threshold $\epsilon$, and its performance does not differ significantly from the model without the loss.
However, by adding $\epsilon$ and setting it to 0.9, the Bridge loss leads to a significant improvement of 0.82 mAP(\%) and 0.56 mAR(\%).
Thus, the final MANet model consists of the Bridge loss, dual-branch structure, and fusion operation, yielding an overall mAP increase of 6.4\% (from 19.97 to 21.25). Qualitative examples of bodily expression estimation using the final model and two baselines are shown in Figure~\ref{fig:result_lma2}.

  \begin{figure*}[ht!]
    \centering
\includegraphics[width=0.75\textwidth]{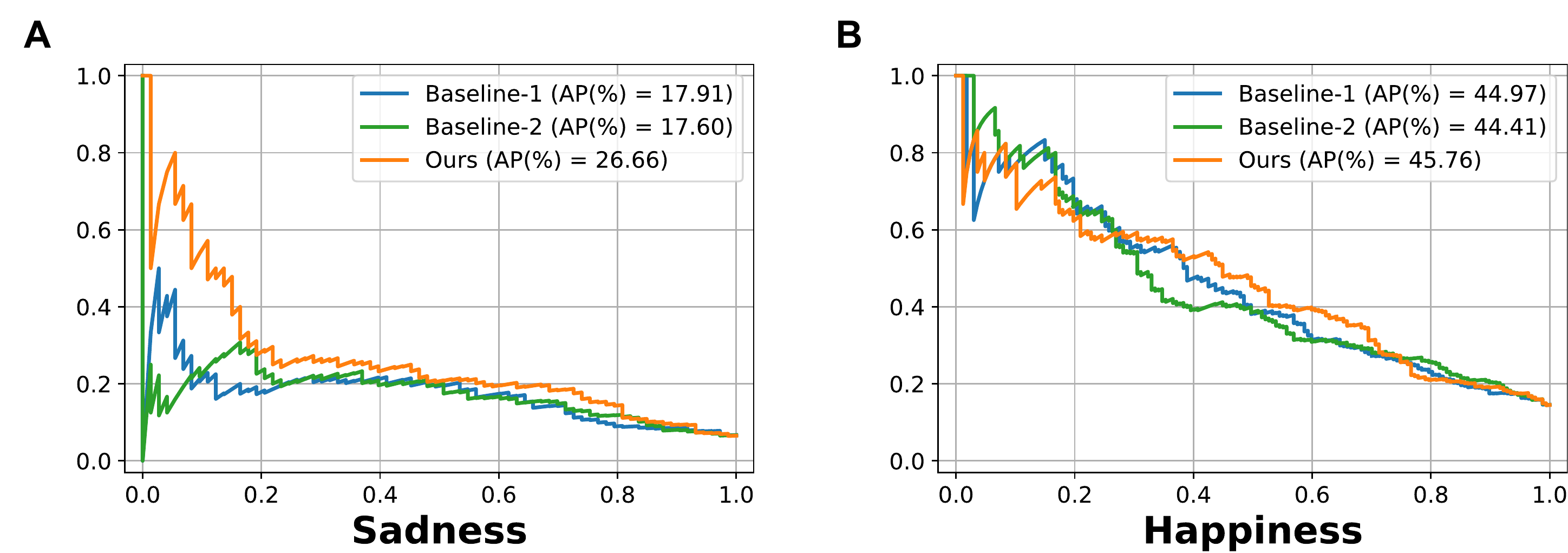}
    \caption{\textbf{Precision-recall curves for sadness and happiness on the BoLD validation set}\hfill\break Models of Baseline-1, Baseline-2 and Ours are identical to those in Figure~\ref{fig:result_lma2}.}
    \label{fig:compare_sad_happy}
    \end{figure*}
    
The core idea of the Bridge loss is to use LMA prediction to supervise the prediction of happiness and sadness.
To further investigate the impact of the Bridge loss on these two emotion categories, we present precision-recall curves for sadness and happiness in Figure~\ref{fig:compare_sad_happy} for the final model and two baselines.
These results show that the Bridge loss leads to a significant improvement in sadness, with an increase of 8.75 AP(\%) and 9.06 AP(\%) over Baseline-1 and Baseline-2, respectively.
There is also a notable improvement in the happiness category.

Table~\ref{tab:sota} compares the performance of MANet with previous state-of-the-art methods.
MANet outperforms the approach by \citeauthor{pikoulis2021leveraging}~\cite{pikoulis2021leveraging} by 1.95 mAP(\%) and 1.38 mAR(\%) on the BoLD validation set. 
On the test set, the single model of MANet achieves comparable performance with model ensembles of \citeauthor{pikoulis2021leveraging}.
Using the same ensemble strategy, the mAP of MANet outperforms the work by \citeauthor{pikoulis2021leveraging}~\cite{pikoulis2021leveraging} by 5.6\%.
The improvement is achieved through the use of BoME as an additional source of training data, despite its smaller size
(approximately one sixth of the BoLD training set).

\begin{table}[ht!]
    \caption{\textbf{Comparison with the state of the art on the BoLD validation and test set}\hfill\break $^*$ represents model ensembles.}\label{tab:sota}
    \centering
    \begin{tabular}{l l cc}
    \toprule
    Set & Method &  mAP(\%) & mAR(\%)   \\
    \hline
    \multirow{5}{*}{Validation} & \citeauthor{luo2020arbee}\citep{{luo2020arbee}}  & 18.55 & 64.27   \\
    & \citeauthor{filntisis2020emotion}\citep{filntisis2020emotion}  & 16.56 & 62.66   \\
    & \citeauthor{pikoulis2021leveraging}\citep{pikoulis2021leveraging}   & 19.30 & 66.94   \\
    & MANet  & \textbf{21.25} & \textbf{68.32}   \\
    \hline
    \multirow{5}{*}{Test} & \citeauthor{luo2020arbee}\citep{luo2020arbee}  & 17.14 & 63.52   \\
    & \citeauthor{filntisis2020emotion}\citep{filntisis2020emotion}  & 17.96 & 64.16   \\
    & \citeauthor{pikoulis2021leveraging}\citep{pikoulis2021leveraging}$^*$   & 21.87 & 68.29   \\
    & MANet  & 22.11 & 67.69  \\
    & MANet$^*$  & \textbf{23.09} & \textbf{69.23}  \\
    \bottomrule
    \end{tabular}
\end{table}

\section{Discussions}\label{section:discussion}
We introduced BoME, a first-of-its-kind LMA-based dataset for human motor elements, and demonstrated the efficacy of deep neural networks in learning human movement representation using this dataset.
We also proposed MANet, a dual-branch model for bodily expressed emotion understanding, which leverages the supervision provided by BoME through a specialized model structure, loss function, and weakly-supervised training strategy to outperform previous BEEU methods. 

This research represents a major step forward in the modeling of bodily expressed emotions. It is well known that different individuals may express the same emotion using different movements. Despite this variability, we are still able to recognize the emotion being expressed by others, even when it is conveyed through movements that differ from those we use ourselves. This is because all motor expressions of a particular emotion are composed of a set of specific motor elements that are associated with that emotion. Once we identify these motor elements, we can recognize the emotion being expressed, regardless of whether we have encountered the specific movements before. For example, anger is characterized by the motor elements of Strong (strength), Sudden (fast), and Direct. Both a kick and a punch may include these motor elements, yet they are distinct movements performed by different body parts. 

Automatic emotion recognition based solely on emotion labels is limited to recognizing emotions from movements that are similar to those included in the training dataset. However, once a machine learns to recognize the relevant motor elements, it can identify them in any type of movement, even those not seen during training. This is why adding motor element labels as input to the emotion recognition process leads to higher recognition rates compared to those obtained from learning based solely on emotion labels.

In this study, we introduce the use of eleven LMA elements known to be related to happiness and sadness to enhance BEEU. This represents a pioneering effort to incorporate the LMA elements into BEEU. While other LMA elements may also be related to emotions, further annotation and labeling efforts by CMAs will allow the scientific community fully explore these relationships. Before carrying out large-scale annotation efforts, however, it is important to determine the relationship between other LMA elements and other emotion categories.
While this work successfully validated the relationship between the eleven elements and happiness and sadness in the BoME and BoLD datasets, the relationship between the remaining LMA elements and other emotion categories remains an area of open research. 
It is possible that future work in psychology or affective computing may provide additional insights into these relationships, ultimately benefiting BEEU research.

Overall, the incorporation of the LMA elements has effectively enhanced BEEU and holds promise for further improvement in the future.

\section{Experimental Procedures}\label{section:experiment}
\subsection{Resource availability}
\subsubsection{Lead contact} 
Request for information and resources used in this article should be addressed to Dr. James Wang\hfill\break (jwang@ist.psu.edu).

\subsubsection{Materials availability}
This study did not generate new unique reagents.

\subsubsection{Data and code availability}
All the codes used in the experiments were implemented with PyTorch. We wrote the code based on the open-source codebase MMaction2.
The BoLD dataset is publicly available at \url{https://cydar.ist.psu.edu/emotionchallenge/index.php}.
We will release the BoME dataset and all the code.

\subsection{LMA-based annotation process}
Each segment of human movement is composed of several motor elements.
Different movements performed by different individuals may share the same set of motor elements.
By recognizing these elements, human movements can be coded.
The relationship between human movement and motor elements is similar to that between music and notes, where a piece of music can be represented as a string of notes, and multiple notes can be played simultaneously, as in a chord.
Several past attempts adopted different sets of movement characteristics to develop distinct human movement coding systems.
For example, \citeauthor{de1989contribution}~\cite{de1989contribution} used the movements in the vertical and sagittal direction, force, velocity, and directness, while 
\citeauthor{montepare1999use}~\cite{montepare1999use} used form, tempo, force, and direction. 
Other researchers~\cite{pollick2001perceiving, sawada2003expression, roether2009critical,gross2010methodology, gross2012effort, barliya2013expression} have employed kinematic variables such as movement duration, velocity, acceleration, and joints displacement.

To characterize motor elements, we adopt the Laban Movement Analysis (LMA), the most extensively developed system for encoding human movement. 
Dance artist, choreographer, and dance theorist Rudolf von Laban (1879-1958) first introduced the LMA in the early 20th century as a means of recording body movement in dance, such as ballet. LMA has four main categories of motor elements.
The \textbf{Body} category lists moving body parts (such as the head and arms) and some typical actions (such as jumping and walking). The \textbf{Space} category represents the body's spatial direction when moving, including vertical (up, down), sagittal (forward, backward) and horizontal (right side/left side).
The \textbf{Shape} category defines how the body changes its shape, including whether it encloses or spreads, rises or sinks. 
The \textbf{Effort} category specifies the inner attitude of the mover toward the movement, comprising four factors--Weight, Space, Time, and Flow.
Weight-Effort refers to the amount of force applied by the mover, with a  
spectrum ranging from Strong (applying high force) to Light (applying weak force) to none (\ie, Passive Weight).
Space-Effort ranges from Direct to Indirect, indicating whether the mover moves directly toward a target in space or indirectly. 
Time-Effort ranges from Sudden to Sustain, denoting the movement's acceleration. 
Flow-Effort ranges from Bound to Free flow, expressing the level of control exerted over the movement.
LMA also includes another category \textbf{Phrasing}, which describes how the motor elements change over time.

Several studies have demonstrated that some LMA elements are strongly associated with emotions.
\citeauthor{shafir2016emotion}~\cite{shafir2016emotion} found that certain LMA elements, when presenting in a movement, can elicit four fundamental emotions (\ie, anger, fear, sadness, and happiness). 
\citeauthor{melzer2019we}~\cite{melzer2019we} identified LMA elements that, when present in a movement, allow it to be classified as expressing one these four fundamental emotions.
The association between happiness and certain LMA elements was also validated by \citeauthor{van2021move}~\citep{van2021move}.
In these studies, the LMA elements that were found to evoke these emotions through movement were largely the same as those used to recognize each emotion through movement.
Our work builds upon these psychological findings related to happiness and sadness. In constructing the dataset, we annotated eleven LMA motor elements that are associated with happiness and sadness (see Table\ref{table:list}). All experiments in this paper are based on these motor elements.

\subsection{The creation of the BoME dataset}\label{sec:data}
To create the BoME dataset, we followed the same process as the BoLD dataset by using movies from the AVA dataset~\cite{gu2018ava} as our data source. This has two advantages. Firstly, real-world video recordings often have limited body movements, but movies provide a rich variety of visual features. Secondly, we can match video clips from the BoLD dataset, which have emotion category labels, for joint training and emotion modeling.

To process the long movies from the AVA dataset into clips, we used the kernel temporal segmentation (KTS) approach~\citep{potapov2014category}, following the time-segmentation in BoLD. The LMA annotator selected the clips based on the following criteria: (1) The annotated human in the clip should have distinct emotions, namely happiness or sadness, as we focus in this study on the eleven motor elements that have been linked to these two emotions in previous studies. (2) The clip should be short, with a total number of frames less than 300, as longer clips may contain expressions of multiple emotions, making it difficult to associate each LMA label with the correct emotion. (3) We excluded clips in which the human does not move at all. We ultimately selected 1,600 clips to form the BoME dataset.

Because each video may contain multiple people, we need to identify which person to annotate.
We followed the process outlined by \citeauthor{luo2020arbee}~\cite{luo2020arbee} for locating the human.
We use the pose estimation network OpenPose~\cite{cao2017realtime} to extract the human joints coordinates, allowing us to obtain a bounding box for the human body. We then use a tracker on the bounding box to assign a unique person ID for the same person in all the frames of the clip.

Next, we ask an LMA expert to provide the annotation. The LMA expert is part of a group of experts who have undergone specialized training in LMA coding for scientific studies, and have coded hours of movements for previous quantitative research using the same standard annotation pipeline used in this study:
First, the annotator must ensure that the sound in the clips is turned off during the annotation so as not to be influenced by what she hears (such as the tone of voice or background music, which could affect the perceived emotion), but only by the movement she sees. 
The annotator should watch each clip as many times as needed to code all eleven variables (\ie, the eleven motor elements that have been shown in previous psychological studies to be associated with motor expressions of sadness and happiness). During each viewing, the annotator codes some of the variables and repeats the process until all variables are coded. The annotator then watches the clip one last time to verify the accuracy of the annotations. If the LMA expert is in doubt regarding the correct coding, she should match what she sees in the clip with her own body movement, and even intensify it when needed, until it becomes clear which motor elements constitute the movement.

The LMA expert codes each motor element with a score, using a standard and consistent scale of 0-4, taking into account both
the duration (\ie, for how long or, in other words, the percentage of the clip duration during which a specific motor element was observed) and the intensity of that observed motor element.
The following standard is used to determine the duration score:
\begin{itemize}
\item 0: The motor element was never observed in that clip.			
\item 1: The motor element was rarely observed, appearing for up to a quarter of the clip duration.
\item 2: The motor element was observed a few times, appearing for up to half of the clip duration.
\item 3: The motor element was often observed, appearing for up to three-quarters of the clip duration.
\item 4: The motor element was observed for most or all of the clip duration.
\end{itemize}
If the intensity of the observed motor element is very low, 1 is subtracted from the duration score. If the intensity is very high, 1 is added to the duration score. However, no score can be higher than 4.

\subsection{Estimating LMA elements on BoME}

The video/action recognition task and LMA elements prediction have a close relationship because they both employ video recordings of humans as input to comprehend human behavior, with one classifying actions and the other recognizing the LMA elements.
We adopt the following four networks to benchmark the BoME dataset. 

\begin{itemize}
    \item TSN~\citep{wang2018temporal} segments videos into several clips, from which we randomly choose one frame.
The extracted frames are fed into the 2D convolutional network, and the prediction is subsequently made using the ensemble findings.
We adopt the popular 2D convolutional network ResNet-101~\cite{he2016deep} as the backbone.
The original TSN is a  two-stream network with optical-flow images and RGB images as input, respectively.
Herein, we only use the RGB input for a fair comparison to other RGB-based methods.
Of note, we crop the region of the human out of the entire RGB image as input, where OpenPose~\cite{cao2017realtime} detects the region.
SlowFast and V-Swin follow this strategy.
    \item SlowFast~\citep{feichtenhofer2019slowfast} models spatiotemporal information using the 3D-convolution operators.
The SlowFast’s 3D-convolutional network has a fast and slow path, where two sampling rates are employed, in order to learn distinct characteristics independently.
We adopt the 3D convoluitonal ResNet-101~\cite{he2016deep} as the slow path.
    \item V-Swin~\citep{liu2022video} changes the well-performing classification network Swin Transformer~\citep{liu2021swin} to a 3D Transformer network. 
We use the base-Swin setting following the original paper.
    \item PoseC3D~\citep{duan2022revisiting} is a skeleton-based method that utilizes a 3D-convolutional network on human poses. 
    The human pose inputs are detected by OpenPose.~\cite{cao2017realtime}
    We adopt the network structure provided by MMaction2.~\cite{2020mmaction2}
\end{itemize}
\subsection{Enhancing emotion understanding with LMA}
We use both the BoME and BoLD datasets to jointly train a model for the recognition of emotion and LMA labels.
As aforementioned, these datasets have a shared set of around 600 video samples. 
The remaining LMA samples only include LMA labels, whereas the remaining BoLD samples only contain emotion labels.
To leverage the LMA information to improve emotion prediction, we designed a dual-branch structure, defined a new loss function, and adopted a weakly supervised approach. 

The framework is illustrated in Figure~\ref{fig:lma}.
The shape of the input is $T\times H \times W \times 3$.
In practice, we sample $T=48$ frames from one video. 
We detect the human region with OpenPose~\cite{cao2017realtime} and then crop and resize the region to $H=224, W=224$ as input.
The backbone follows the first three stages of V-Swin~\cite{liu2022video} precisely.
Each stage comprises numerous 3D Swin Transformer blocks, where the V-Swin paper~\cite{liu2022video} describes those blocks' construction.
We adopt the base setting of V-Swin, which consists of 2, 2, and 18 blocks in the first three stages, respectively.
Following the backbone, we add two Swin Transformer blocks as the emotion branch and another two blocks as the LMA branch.
Of note, before the training, we directly use the Kinect-400 pre-trained V-Swin to initialize the backbone and the emotion branch.

The output of the emotion branch is $y=\{y_i\}^{N}_{i=1}$, where $N$ is the number of emotion categories.
Similarly, $z=\{z_j\}^{M}_{j=1}$ is the output of the LMA branch, where $M$ is the number of the LMA elements. 
In BoME and BoLD, we have $N=26, M=10$.
Let us denote the ground truth emotion and LMA label as $\hat{y}$ and $\hat{z}$.
Considering the LMA and emotion prediction are both the multi-label binary classification problem, we compute the multi-label cross-entropy loss:
\begin{equation*}
\begin{split}
    \mathcal{L}_\text{Emotion} & = -\frac{1}{N} \sum^{N}_{i=1}\hat{y}_i\ln\sigma (y_i)+ (1-\hat{y}_i)\ln(1-\sigma (y_i))\;, \\
    \mathcal{L}_\text{LMA}& = -\frac{1}{M} \sum^{M}_{j=1}\hat{z}_j\ln\sigma (z_j)+ (1-\hat{z}_j)\ln(1-\sigma (z_j))\;,
\end{split}
\end{equation*}
where $\sigma$ is the sigmoid function.

Furthermore, using the knowledge that these LMA elements are highly related to sadness or happiness, we design the Bridge loss to supervise the sadness and happiness prediction.
We pick the maximum prediction from the sadness-related and happiness-related LMA elements and then compute the softmax of them to get the probability for happiness and sadness. 
Formally, we have,
\begin{equation*}
    \begin{split}
        p_\text{happiness}= \frac{e^{max\{z_i\}^{M}_{i=M^*+1}}}{e^{max\{z_i\}^{M*}_{i=1}} + e^{max\{z_i\}^{M}_{i=M^*+1}}}\;, \\
        p_\text{sadness}= \frac{e^{max\{z_i\}^{M^*}_{i=1}}}{e^{max\{z_i\}^{M*}_{i=1}} + e^{max\{z_i\}^{M}_{i=M^*+1}}}\;,
    \end{split}
    \end{equation*}
where the first $M^*$ elements of $z$ are related to sadness and the others are related to happiness. In BoME, $M^*=4$.
$p_\text{happiness}$ and $p_\text{sadness}$ are the probabilities from the perspective of LMA prediction. Then we can use them to supervise the happiness and sadness probability derived from the emotion output via the soft cross-entropy loss:
\begin{equation*}
    \mathcal{L}_\text{Bridge} = -p_\text{happiness}\ln\frac{e^{y_h}}{e^{y_h}+e^{y_s}}-p_\text{sadness}\ln\frac{e^{y_s}}{e^{y_h}+e^{y_s}}\;,
\end{equation*}
where $y_h$ and $y_s$ are the happiness and sadness output of $y$.
However, the happiness-sadness prediction from the LMA branch is not accurate sometimes,  preventing it from supervising $y_h$ and $y_s$.
Thus, we set a threshold $\epsilon$ - only when $p_\text{happiness}$ or $p_\text{sadness}$ is greater than $\epsilon$, we will compute the cross entropy loss. Formally, we have
\begin{eqnarray*}
    \mathcal{L}_\text{Bridge} &=& -\mathbb{1}(p_\text{happiness}>\epsilon)\ln\frac{e^{y_h}}{e^{y_h}+e^{y_s}}\\ &&- \mathbb{1}(p_\text{sadness}>\epsilon) \ln\frac{e^{y_s}}{e^{y_h}+e^{y_s}}\;.
\end{eqnarray*}
We perform the ablation study on the different $\mathcal{L}_\text{Bridge}$ in Table~\ref{tab:loss}. The results indicate that the $\epsilon$-controlled loss function can get better performance. Specifically, $\epsilon=0.9$ achieves the best performance.

Inspired by \citeauthor{wu2020mebow}~\cite{wu2020mebow}, we adopt a weakly supervised training strategy to leverage data that do not have both emotion and LMA labels. 
We use the coefficient $\mu_\text{Emotion}$ with 0 or 1 to indicate whether this sample has the emotion label or not. 
So does the coefficient $\mu_\text{LMA}$.
In the training, we mix and shuffle all the data sets together.
We also use the coefficients $\lambda_1, \lambda_2$ to balance the three losses.
The total loss is,
\begin{equation*}
    \mathcal{L} = \mu_\text{Emotion} \mathcal{L}_\text{Emotion} + \lambda_1 \mu_\text{LMA} \mathcal{L}_\text{LMA} + \lambda_2 \mathcal{L}_\text{Bridge}\;,
\end{equation*}
where we set $\lambda_1=0.25$ and $\lambda_2=0.1$ in practice.

During the training stage, the network undergoes 50 epochs of training. Data augmentation techniques flipping and scaling are applied to the BoME and BoLD datasets.
The learning rate is set at 5e-3 and the optimization algorithm used is SGD.

\section{Acknowledgments} 
The work was supported in part by generous gifts from the Amazon Research Awards program to J. Wang. This work used cluster computers at the Pittsburgh Supercomputer Center through an allocation from the Advanced Cyberinfrastructure Coordination Ecosystem: Services \& Support (ACCESS) program, which is supported by National Science Foundation (NSF) grants Nos. 2138259, 2138286, 2138307, 2137603, and 2138296. This work also used the Extreme Science and Engineering Discovery Environment, which was supported by NSF grant No. 1548562. We are grateful to Laura Schandelmeier for providing the LMA annotations for BoME and to Yelin Kim and Adam Fineberg for their support and encouragement.

\section{Author Contributions}

C.W. developed the deep learning pipeline, analyzed the data, wrote the manuscript, and contributed to the data curation. D.D. analyzed the data and contributed to the writing. T.S. and R.T. contributed to the LMA labeling process and commented on the manuscript. J.Z.W. served as the principal investigator of the project, engaged in technical discussion, and provided extensive editing and comments on the manuscript.

\section{Declaration of Interests}
The authors declare no competing interests. 

\bibliographystyle{model3-num-names}
\bibliography{bib}
\end{document}